\documentclass[letterpaper, 10 pt, final, conference ]{ieeeconf}  

\IEEEoverridecommandlockouts                              

\usepackage[dvipsnames]{xcolor}

\usepackage{graphics} 
\graphicspath{{img/}}
\usepackage{epsfig} 
\usepackage{amsmath} 
\usepackage{amssymb}  
\usepackage{pifont}
\usepackage[makeroom]{cancel}
\usepackage{xcolor} 

\usepackage[center]{subfigure}
\usepackage{multirow}
\usepackage{booktabs}
\usepackage{subfigure}
\usepackage{epstopdf}
\usepackage{algpseudocode,algorithm} 

\usepackage{textcomp}

\makeatletter
\let\NAT@parse\undefined
\makeatother
\usepackage{hyperref}  
\hypersetup{
  colorlinks,
  citecolor=Violet,
  linkcolor=Green,
  urlcolor=Blue
  }

\title{\LARGE \bf Adaptive robot body learning and estimation through predictive coding}

\author{Pablo Lanillos, Gordon Cheng
\thanks{All authors are affiliated to the Institute for Cognitive Systems (ICS),  Technische Universit\"at M\"unchen, Institute for Cognitive Systems, Arcisstraße 21 80333 M\"unchen, Germany
		{\tt\small \{p.lanillos, gordon\}@tum.de}.}
\thanks{This work was supported by SELFCEPTION project (www.selfception.eu) European Union Horizon 2020 Programme (MSCA-IF-2016) under grant agreement no. 741941.}
\thanks{Preprint version. Accepted for publication in IEEE International Conference on Intelligent Robots and Systems (IROS 2018).}
}

\begin{document}

\maketitle
\thispagestyle{empty}
\pagestyle{empty}

\begin{abstract}

The predictive functions that permit humans to infer their body state by sensorimotor integration are critical to perform safe interaction in complex environments. These functions are adaptive and robust to non-linear actuators and noisy sensory information. This paper introduces a computational perceptual model based on predictive processing that enables any multisensory robot to learn, infer and update its body configuration when using arbitrary sensors with Gaussian additive noise. The proposed method integrates different sources of information (tactile, visual and proprioceptive) to drive the robot belief to its current body configuration. The motivation is to enable robots with the embodied perception needed for self-calibration and safe physical human-robot interaction.

We formulate body learning as obtaining the forward model that encodes the sensor values depending on the body variables, and we solve it by Gaussian process regression. We model body estimation as minimizing the discrepancy between the robot body configuration belief and the observed posterior. We minimize the variational free energy using the sensory prediction errors (sensed vs expected). 

In order to evaluate the model we test it on a real multisensory robotic arm. We show how different sensor modalities contributions, included as additive errors, improve the refinement of the body estimation and how the system adapts itself to provide the most plausible solution even when injecting strong sensory visuo-tactile perturbations. We further analyse the reliability of the model when different sensor modalities are disabled. This provides grounded evidence about the correctness of the perceptual model and shows how the robot estimates and adjusts its body configuration just by means of sensory information.

\end{abstract}



\section{Introduction}
\label{sec:intro}
Providing the robot with the predictive functions of the body, environment and others is a critical aspect for complex interaction. Appropriately, in order to generate safe and robust interaction the artificial agent must take into account uncertainties related to the sensory input as well as the unexpected events that can occur. Unfortunately, a perfect model of the body, environment and others is almost impossible to design. We present an adaptive robot body learning and estimation algorithm able to deal with noisy sensory inputs and to integrate multiple sources of information (touch, visual and proprioceptive sensors). This model is framed on the predictive processing theory proposed by Friston \cite{friston2005theory} and biologically grounded on predictive coding evidence about the brain as observed by Rao and Ballard in the visual cortex \cite{rao1999predictive}. This approach has been extensively studied in computational biology and psychology but it has not been properly tested in robotics \cite{pio2016active}.

\begin{figure}[t!]
\centerline{\includegraphics[width=0.95\columnwidth]{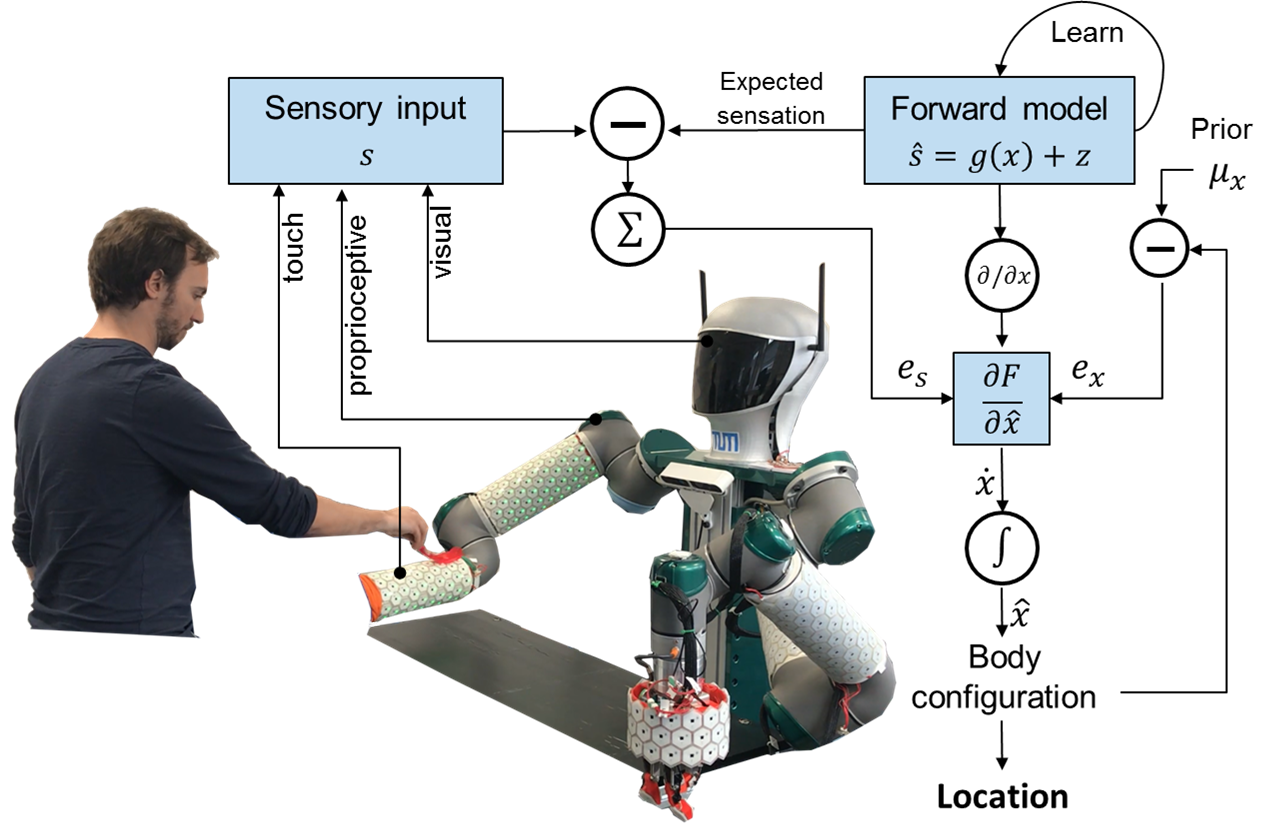}}
\caption{Proposed adaptive robot body learning and estimation using prediction errors: expected sensation minus sensory input. Visual, tactile and proprioceptive sensing contribute to obtain the most plausible body configuration, and hence the end-effector location. Body inference is computed by minimizing the free-energy with respect to the latent variables.}
\label{fig:abstract}
\end{figure}

The main idea behind this embodied approach of robot body perception is that the only available information is the sensory input \cite{lanillos2016yielding}. By learning the predictors of the sensor outcome given its current body latent variables and the actions exerted, the robot is able to properly infer its real body configuration. The error between the expected sensory signal and the real input contributes to refine the most plausible hypothesis that the robot has about its body, as depicted in Fig. \ref{fig:abstract}. This simplifies the complexity of online estimation of the body internal variables and increases the ability of the robot to adapt to uncertain situations. 



\subsection{Motivation and method}
To produce safe interaction the robot should robustly predict its body and other agents in every instant using all sensory information available. This undoubtedly passes through the systematic design of the body model or enabling the robot with an accurate perception of its body \cite{lanillos2017enactive}. Here, we define body perception from the probabilistic perspective \cite{thrun2005probabilistic} as inferring the body variables $x$ only depending on the sensory information $s$: $p(x|s)$\footnote{We have intentionally left the action out to be coherent with the active inference perceptual theory from Friston. At the end of the paper we remark the role of the action within the proposed scheme.} \cite{lanillos2016yielding}. When the robot does not have access to its body variables we can infer the body configuration from the sensory input through Bayes:



\small
\begin{equation}
p(x|s)= \frac{p(s|x)p(x)}{p(s)}
\end{equation}
\normalsize
where $p(s|x)$ is the sensory consequence of being in state $x$ and $p(x)$ is the prior belief of the internal variables. We could estimate this posterior using Bayesian recursive filters \cite{thrun2005probabilistic}. However, instead of computing this posterior directly, we approximate an auxiliary distribution over the unobserved latent variables $q(x)$ to the real posterior $p(x|s)$. Moreover, according to predictive processing theory \cite{friston2008hierarchical}, the robot \textit{belief}, and variables that represent it, differs from the real world process\footnote{We describe the robot mind as a system whose belief has its own dynamics and internal variables and the inference process tries to fit the robot internal representation to the real world.}. Moreover, it has incomplete knowledge about the real generative process of its body. Thus, we have to approximate, at every instant, not only the state variables but a particular density of a family of functions. In other words, we approximate the distribution $q(r)$ where $r$ is the internal state of the robot. For this paper, as the body configuration has been simplified to the joint angles, we overloaded the notation by using $x$ for the body state and $r = \hat{x}$ as the internal state of the robot.

In order to approximate both distributions (real and believed) we can minimise the Kullback-Leibler divergence $D_{KL}(q(\hat{x}) || p(x|s))$, through the free-energy $F$ bound, expressed as \cite{friston2008hierarchical,buckley2017free}:

\small
\begin{equation}
\label{eq:KL_F}
D_{KL}(q(\hat{x}) || p(x|s)) = F + \ln p(s)
\end{equation}
\normalsize
When $F$ converge to $\emptyset$, only the sensory surprise differs from the belief and the posterior is properly approximated. Hence, in theory, the inference of the body internal configuration depending on the sensory input can be tractable approximated by minimizing the variational free energy. One way to minimize it is through a gradient descent scheme: $\dot{x} = \frac{\partial F}{\partial x}$.

From the generative process point of view, $\mathbf{f}(x)$ governs the dynamics of the environment and $\mathbf{g}(x)$ governs the sensory information. However, the robot has an approximation of them $f(\hat{x})$, $g(\hat{x})$. Thus, the agent is continuously adapting its belief about its body and the world just with the sensory input. This is performed by dynamically updating its internal variables $\hat{x}$ by means of
the error between the expected and the real sensory input: \textit{the prediction error}.

Conversely to previous works on predictive coding, instead of knowing the sensor generative functions $g(x)$ here we provide a method to learn them and transparently integrate them into a predictive coding scheme. 


\subsection{Related works}
\label{sec:background}
Predictive coding \cite{rao1999predictive} and predictive processing \cite{friston2005theory}, has mainly been studied for human perception and control. Just a few works have applied it to robots. For instance, in \cite{pio2016active} the viability of predictive processing for robot control on a simulated robotic arm is discussed. However, the generative functions and the parameters were known in advance. Besides, implementing predictive coding with deep neural networks has gained popularity for modelling multisensory perception \cite{ahmadi2017bridging} and, in the computer vision community, for video prediction \cite{lotter2016deep}. Finally, this approach shares some conceptual background with sensorimotor contingencies \cite{lanillos2016yielding,angulo2017dynamical} approaches, and predictive learning \cite{nagai2015predictive}, where the robot develops its perception as infants do.

The variational approach presented in this paper is related to expectation-maximization algorithms formalized as a maximization-maximization problem of the free-energy function \cite{neal1998view}. In fact, using Friston terminology, predictive processing is a dynamic expectation-maximization algorithm \cite{friston2008hierarchical}. Furthermore, it is important to highlight the strong similarities with the ensemble Kalman filter \cite{evensen1994sequential}. We have adopted the free-energy mathematical framework for the following reasons: it provides indirect minimization of the Kullback-Leibler divergence \cite{friston2008hierarchical}; it supports multisensory non-linear integration; it is scalable in its Laplace approximation \cite{buckley2017free}; it permits unsupervised parameters tuning \cite{bogacz2015tutorial}; and it is biologically plausible \cite{rao1999predictive}.

In terms of body model learning there is a vast literature on regressors such as locally weighted projection regression \cite{vijayakumar2005incremental}, local Gaussian process \cite{nguyen2009local} or infinite experts algorithm \cite{damas2012online}. These methods are able to compute the mapping between the sensory input and the configuration of the body and are used to learn forward and inverse kinematics and dynamics. Feed-forward and recurrent nets can also be assimilated for learning body schemas and the needed predictors but relay on supervised information and hundreds of parameters optimization \cite{nabeshima2006adaptive,wieser2016progressive}. Unsupervised and self-exploration learning of the body has also been addressed in works like \cite{stoytchev2011self} using temporal contingencies. Moreover, biologically plausible sensorimotor learning has been investigated in works like \cite{mori2010human} by means of Hebbian-based methods where body calibration can be learnt through sensorimotor mapping. Dynamic Hebbian learning has also been proposed for obtaining intermodal forward models in \cite{schillaci2016exploration}. Body model free visual detection ~\cite{lanillos2016self} has been approached as an intermodal inference problem but it is restricted to the camera view of the robot.



\subsection{Contribution and organization}
This work introduces predictive processing for robot body perception \cite{friston2008hierarchical}, where the robot first learns the sensors or features forward generative models and then it is able to dynamically provide the most plausible body configuration and the location of the end-effector, incorporating in a scalable way several noisy sources of sensory information. We address free model body learning and estimation, where the sensor generative/forward model is learnt using a Gaussian process regression. The body configuration and end-effector location is obtained by means of on-line free-energy minimization using the prediction error.

The computational model is presented in Sec. \ref{sec:model}, where we propose a way to learn the generative sensor model $g(x)$ and its derivative by exploration (sampling), as well as the differential equations to solve body estimation through free-energy minimization. The experimental set-up on a real multisensory robot is presented in Sec. \ref{sec:experiment}. In Sec. \ref{sec:results} we analyse the proposed approach, evaluating the body estimation with different sensor modalities and inducing visuo-tactile perturbations. Finally, in Sec. \ref{sec:discussion} and \ref{sec:conclusion} we discuss the advantages and drawbacks of the proposed approach, and enforce the applicability of the method to improve self localization and interaction.

\section{Mathematical model}
\label{sec:model}
\begin{table}[htbp!]
\centering 
\resizebox{\columnwidth}{!}{
\begin{tabular}{r p{5cm}}
\textbf{Notation} & \\
$\mathbf{x}, x$ &  Distribution and value of the body vars. \\
$\hat{x}$ & Most plausible hypothesis of body vars.\\
$\mu_x$ & Prior belief of the body variables\\
$s: s_v, s_p, s_t$ & Sensor value: visual, proprioceptive, tactile\\
$e: e_v, e_p, e_t$ & Error value: visual, proprioceptive, tactile\\
$F$ & Free-energy \\
$\frac{\partial F}{\partial x} $ & Derivative of free-energy w.r.t. internal state \\
$f(s; g(.), \sigma) $ & Normal with $g(.)$ mean and $\sigma$ variance \\
\end{tabular}
}
\label{table:notation}
\end{table}

We first describe the proposed mathematical model for visual and proprioceptive information and then we extend it with a more complex visuo-tactile input. The model is based on works from predictive processing \cite{friston2008hierarchical} and free-energy approaches to perception \cite{bogacz2015tutorial,buckley2017free}. For this model and without loss of generality, we restrict that the robot cannot perceive the gradient of the sensor signal and that the state transition model (generative function $f(\hat{x}$) is believed to be static. In Sec. \ref{sec:results} we discuss the drawbacks of these simplifications. For the sake of clarity, we adopted the free-energy derivation presented in \cite{bogacz2015tutorial}, although the original one uses the KL-divergence as the starting point.

The robot is defined as a set of sensors $s$ and body internal unobserved variables $\hat{x}$. The proprioceptive sensors $s_p$ outputs a value depending on the body configuration that follows a Normal distribution with linear or non-linear mean $g_p(x)$: $p(s_p|x) = f(x; g_p(x), \sigma_p)$. The visual sensor $s_v$ provides the location of the end-effector in the visual field also following a Normal distribution with linear or non-linear mean $g_v(x)$: $p(s_v|x)= f(x; g_v(x), \sigma_v$). Finally, the robot counts with artificial skin sensors on the end-effector limb and it is able to detect other's hand in the visual field - See Fig. \ref{fig:abstract}. 

\subsection{Perception model for visual and proprioceptive sensors}

The body configuration $\mathbf{x}$ can inferred via visual and proprioceptive sensory information through a Bayes rule. Assuming that the visual and proprioceptive sensing are independent, the distribution of $x$ is:
\begin{equation}
p(\mathbf{x}|s_p,s_v) = \frac{p(s_p|\mathbf{x})p(s_v|\mathbf{x}) p(\mathbf{x})}{p(s_v,s_p)}
\end{equation}
The denominator $p(s_v,s_p)$ has integrals that make intractable exact computation for large distributions. 

As explained previously we want to approximate the \textit{belief} distribution $q(\hat{x})$ to the posterior $p(\mathbf{x}|s_p,s_v)$ using Eq. \ref{eq:KL_F}. Mimicking predictive processing theory, which states that the brain works with the most plausible model of the world to perform predictions, instead of working with the whole distribution $\mathbf{x}$, we use the most plausible value: $\hat{x}$. This have an important implication as the denominator does not any more depend on $\hat{x}$ \cite{bogacz2015tutorial}, and hence we get\footnote{In the ensemble Bayesian filtering terminology this is similar to maintain a sample drawn from the latent space distribution.}:

\begin{equation}
p(\hat{x}|s_p,s_v) = p(s_p|\hat{x})p(s_v|\hat{x}) p(\hat{x})
\end{equation}
Applying logarithms we obtain the negative free-energy formulation:
\begin{equation}
F = \ln p(s_p|\hat{x}) + \ln p(s_v|\hat{x}) + \ln p(\hat{x})
\end{equation}
Substituting the probability distributions by their functions $f(.;.)$, and under the Laplace approximation \cite{friston2008hierarchical,buckley2017free} and assuming normally distributed noise, we can compute the negative free energy as: 
\small
\begin{align}
\label{eq:freeenergyexample}
F &=   \ln f(s_p; g_p(\hat{x}), \sigma_p) + \ln f(s_p; g_v(\hat{x}), \sigma_v) + \ln f(x; \mu_x, \sigma_x) \nonumber\\
  &=   - \frac{(s_p - g_p(\hat{x}))^2}{2\sigma_p}   -\frac{(s_v - g_v(\hat{x}))^2}{2\sigma_v}   - \frac{(\hat{x} - \mu_x)^2}{2\sigma_x} + \nonumber\\
  & \quad + \frac{1}{2} \left[  -\ln \sigma_x  - \ln \sigma_{s_p} - \ln \sigma_{s_v}  \right] + C. 
\end{align}
\normalsize
To approximate the posterior distribution we minimize $F$, following a gradient-descent scheme:

\begin{equation}
\dot{\hat{x}} = \frac{ \partial F } {\partial \hat{x}}
\end{equation}
Computing the partial derivative of Eq. \ref{eq:freeenergyexample} we obtain:
\begin{align}
\label{eq:dotx1}
\dot{\hat{x}} = \underbrace{- \frac{\hat{x}-\mu_x}{\sigma_x}}_{\text{Error prior}}  + \underbrace{\frac{s_p - g_p(\hat{x})}{\sigma_p}}_{\text{Error proprio.}} g_p'(\hat{x}) +\underbrace{\frac{s_v - g_v(\hat{x})}{\sigma_v}}_{\text{Error visual}} g_v'(\hat{x})
\end{align}

Note that the first term is the error between the most plausible value of the body configuration and its prior belief ($e_x$), the second term is the error between the observed proprioceptive value and the expected one ($e_p$) and the third term is the prediction error between the visual sensed position of the end-effector and the expected location ($e_v$). In order to use Eq. \ref{eq:dotx1} we need to know or learn the sensor forward/generative functions.

For the sake of simplification, we encode the internal state directly in the proprioceptive sensing space, thus defining body just by means of the proprioception state. For that purpose, we substitute $g_p(\hat{x})$ by $\hat{x}$ and its partial derivative $g_p'(\hat{x})$ is set to 1. In other words, if the body configuration is defined by the joint angles, the state $\hat{x}$ will represent the joint sensors (encoders) output. For notation convenience we maintain $\hat{x}$ as the body configuration but it represents $\hat{s}_p$. By defining prediction errors as:



\begin{align}
& e_x = \frac{1}{\sigma_x} (\hat{x}-\mu_x)\\
& e_p = \frac{1}{\sigma_p} (s_p - \hat{x})\\
& e_v = \frac{1}{\sigma_v} (s_v - g_v(\hat{x}))
\end{align}
We construct the differential equation that infers the body latent variables $\hat{x}$ as\footnote{Note that the computation of $\hat{x}$ is a simplification of the predictive processing approach for passive static perception \cite{bogacz2015tutorial} as we are omitting the generative model of the world $f(x)$. Accordingly, to certainly reduce the difference between the believed distribution and the observed one, $e_x$ should describe the error between the world generative function and the internal belief: $f(x) - \hat{x}$. We leave this extension for further works and in Sec. \ref{sec:conclusion} we point out the challenges to obtain the full construct without knowing $f(x)$.}:

\begin{align}
\label{eq:errordotx}
\dot{\hat{x}} = -e_x + e_p + e_v g_v'(\hat{x})
\end{align}

According to Eq. \ref{eq:errordotx} the update of the internal state is driven by the observed and the expected value of the state and the error prediction. The gradient or Jacobian of the sensor with respect to the latent variables maps the contribution of each sensor modality to each body configuration variable in the same way as in the extended or ensemble Kalman filter.

Generalizing the free energy minimization for $i$ sensors the body configuration is driven by:
\begin{align}
\label{eq:partialF1}
\dot{\hat{x}} = - e_x + e_p +\sum_i\frac{\partial{g_i(\hat{x})^T}}{\partial{\hat{x}}} e_i
\end{align}
Then, the full dynamics of our body estimation model is given by:
\begin{align}
&\dot{x} = -e_x + e_p  + e_v g_v'(\hat{x}) \nonumber\\
&\dot{e}_x = s_x  -  \mu_x  - \sigma_x e_x \nonumber\\
&\dot{e}_p = s_p  -  \hat{x} - \sigma_p e_p\nonumber\\
&\dot{e}_v = s_v  -  g_v(\hat{x}) - \sigma_v e_v\nonumber\\
&\dot{\mu}_x = \mu_x + \lambda e_x
\label{eq:bodyestimation}
\end{align}
where $\lambda$ is the learning ratio parameter that specifies how fast the prior of body configuration $\mu_x$ is adjusted to the prediction error. 

\subsection{Body learning -- learning the sensory states caused by the body configuration}
\label{sec:model:learn}
We define body-learning as obtaining the unknown forward/observation model $s= g(x)$ and its derivative/Jacobian $g'(x)$ that relate the sensor values with the body state. This is a consequence of describing body estimation by means of Eq. \ref{eq:bodyestimation}. To learn both functions we use Gaussian process regression with collected data generated by body exploration. We obtain sensor samples $\overline{s}$ from the robot in several body configurations $\overline{x}$. For instance, for the visual generative process $x$ is the proprioceptive state and $s$ is the visual information.


The \textit{training} is performed by computing the covariance matrix $K(X,X)$ on the collected data with noise $\sigma_n^2$, where the covariance function $k(x_i,x_j)$ is defined as:
\begin{equation}
\label{eq:covariance}
k_{ij} = \sigma_n^2 + k(x_i,x_j) \quad \text{where} x_i,x_j \in \overline{x}
\end{equation}
The \textit{prediction} of the sensory outcome $s$ given $x$ is then computed as \cite{rasmussen2005GPM}:
\begin{equation}
\label{eq:GPmean}
 g(\hat{x}) = k(\hat{x},X) K(X,X)^{-1} \overline{s}= k(\hat{x},X) \boldsymbol{\alpha}
\end{equation}
where $\boldsymbol{\alpha} = \text{choleski}(K)^T \backslash ( \text{choleski}(K) \backslash \overline{s})$ for numerical stability.

Finally, in order to compute the gradient of the posterior $g(x)'$ we differentiate the kernel \cite{mchutchon2013differentiating}, and obtain its prediction analogously as Eq. \ref{eq:GPmean}:
\begin{align}
 g(\hat{x})' &= \frac{\partial k(\hat{x},X)}{\partial \hat{x}} K(X,X)^{-1} \overline{s} = \frac{\partial k(\hat{x},X)}{\partial \hat{x}} \boldsymbol{\alpha}
\end{align}
Using the squared exponential kernel with the Mahalanobis distance covariance function, the derivative becomes:
\begin{align}
\label{eq:gderivative}
g(\hat{x})' = -\Lambda^{-1} (\hat{x} - X)^T (k(\hat{x},X)^T \cdot \boldsymbol{\alpha})
\end{align}
where $\Lambda$ is a matrix where the diagonal is populated with the length scale for each dimension ($\text{diag}(1/l^2)$) and $\cdot$ is element-wise multiplication.


\subsection{Adding tactile feedback and other's interaction}
We exploit the artificial skin of the robot to refine the body-configuration estimation. For that purpose, we model the intermodal relation between visual and the tactile sensing\cite{lanillos2016yielding}. When somebody touches the robot end-effector, it should adjust its body configuration to fit the end-effector location in the visual field where the other agent is touching. In other words, other agent touching the robot end-effector in $o_v$ location in the visual field $\rightarrow$ robot end-effector $s_v$ is there $\rightarrow$ body configuration is adjusted.

First, we assume that the robot is able to discern that its end-effector limb is being touched, and that it knows the relation between the touch signal and the location on the body. We define the likelihood function of being touch by other $p(\tau, o_v|x)$ by means of spatial $f_s$ and temporal $f_t$ coherence. We can learn this function by touching the limb in different end-effector locations. Alternatively, in this paper we reuse learnt model from the visual field $g_v$ to compute the expected end-effector location and define the visuo-tactile sensory likelihood as:
\begin{align}
\label{eq:touch}
g_t(\hat{x}) = f_s \cdot f_t = a_1 e^{-b_1\sum(g_v(x) - o_v)^2} \cdot a_2e^{-b_2 \delta^2}
\end{align}
where $a_1,b_1,a_2,b_2$ are parameters that shape the likelihood and have been tuned in concordance with the data acquired in \cite{samad2015perception} from human participants; $\delta$ is the level of synchrony of the event (e.g., time difference between the visual and the tactile event); and $o_v$ is the other agent end-effector location in the visual field.

\begin{figure*}[!t]
\centering
\subfigure[Sensory data acquisition]{\includegraphics[width=0.65\textwidth, height=90px]{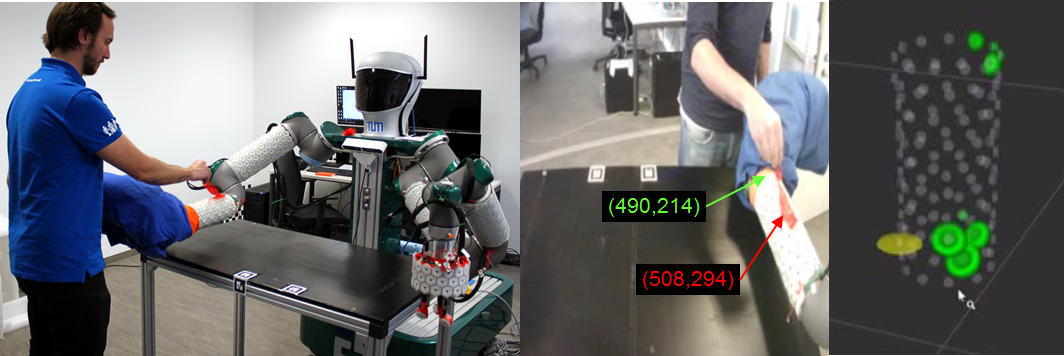} 
\label{fig:setup:a1}}
\subfigure[Adaptation test]{\includegraphics[width=0.30\textwidth, height=90px]{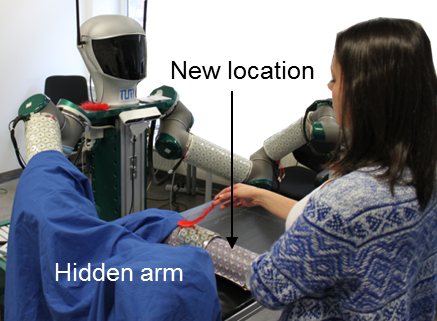}
\label{fig:setup:a2}}
\caption{Experimental setup. (a) Gathering proprioceptive (joint angles), visual robot (green) and other (red) end-effector pixel coordinates) and tactile sensory data from the robot (proximity values) with different participants and positions.  Green circles represent the likelihood of being touched. (b) Adaptation test where we change the visual location of the arm and we induce synchronous visuo-tactile inputs.}
\label{fig:setup}
\end{figure*}

We directly introduce this generative function into the free-energy scheme as follows\footnote{Under the predictive processing framework we might include another internal variable that defines being touched and a second layer of hierarchy that is able to infer similarity (temporal and spatial) between the patterns generated in the visual field by the other agent and the patterns perceived in the skin.}:

\small
\begin{align}
\label{eq:xwithtouch}
\dot{x} =   - \frac{\hat{x}-\mu_x}{\sigma_x} + \left[\frac{s_p - \hat{x}}{\sigma_p}, \frac{s_v - g_v(\hat{x})}{\sigma_v}, \frac{s_t - g_t(\hat{x})}{\sigma_t} \right] \left[\begin{array}{c} \mathbf{1}\\
g_v'(\hat{x})\\
g_t'(\hat{x})
\end{array}\right]
\end{align}
\normalsize
When a synchronous touching and visual pattern occurs the body configuration is adjusted depending on the expected end-effector visual location $g_v(\hat{x})$ and the other's visual location $o_v$.

\subsection{ Adaptive body estimation and learning through predictive processing }

Algorithm \ref{alg:estimation} summarizes the learning and estimation stages to dynamically compute the internal body configuration based on the sensory error prediction, using for multiple independent sources $i \in N$ of sensory information or features, and body internal variables $j \in M$. The \textit{learning} stage is using GP regression described in \cite{rasmussen2005GPM} for each sensor modality contribution to the body configuration. Using the $\boldsymbol{\alpha}\text{-GP}$ solution we reduce the complexity of the prediction calculation. The \textit{estimation} stage computes the prediction error for every sensor and solves the differential equations by variational free-energy minimization. Note that we are applying 1st order Euler integration method. More accurate approaches are out of the scope of this paper.


%

\begin{algorithm}[!hbtp]
\centering
\caption{Multisensory body learning and estimation}
\label{alg:estimation}
\begin{algorithmic}
\small
\State \textbf{\textit{Forward sensor model learning using GP}} 
\Require $X,\overline{s}_i$ \Comment{samples $\overline{s}=g(X)$}
\State	$K =  \text{choleski}(\sigma_n^2\mathbf{I} + k(X,X)$) \Comment{Covariance}
\For {i=1:N} \Comment{For every sensor/feature modality}
\State	$\boldsymbol{\alpha}_i = K^T \backslash ( K \backslash \overline{s})$
\EndFor	
\State \Return $X, \boldsymbol{\alpha}_i$
\\\noindent\rule{\columnwidth}{0.6pt}
\State \textbf{\textit{Predictive processing body estimation}}
\Require $X, \boldsymbol{\alpha}_i $ \Comment{GP training}
\Require $\mu_x \in \mathbb{R}^M$ \Comment{Body configuration prior}
\Require $\hat{x} \in \mathbb{R}^M$ \Comment{Initial body estimation}
\Require $e \in \mathbb{R}^N$ \Comment{Initial predition error}
\State $s_i \gets \mathbf{g}(x)$ \Comment{Input sensor information}
\For {i=1:N} \Comment{Compute $N$ predictions}
\State $g_i(\hat{x}) = k(\hat{x},X) \boldsymbol{\alpha}_i$ \Comment{Eq. \ref{eq:GPmean}}
\State $g'_i(\hat{x}) = -\Lambda^{-1} (\hat{x} - X)^T (k(\hat{x},X)^T \cdot \boldsymbol{\alpha}_i)$ \Comment{Eq. \ref{eq:gderivative}}

\EndFor

\For {i=1:N} 
\State $\dot{e}_i = s_i  -  g_i(\hat{x}) - \sigma_i e_i$ \Comment{Prediction errors Eq. \ref{eq:xwithtouch}}
\EndFor
\State $\dot{\hat{x}} = -e_x + \sum_i e_i g_i'(\hat{x})$ \Comment{Free-energy minimization dynamics}
\State $\dot{e}_x = \hat{x}  -  \mu_x  - \sigma_x e_x $ 
\State $\dot{\mu}_x = \mu_x + \lambda\ e_x$
\State $\mathcal{X} \gets \left[\hat{x},e_{1:N},e_x,\mu_x\right]^T$
\State $\mathcal{X} = \mathcal{X} + \Delta_t \mathcal{\dot{X}}$ \Comment{Integration}


\State \Return $\mathcal{X}$

\end{algorithmic}
\end{algorithm}

\normalsize

\section{Experimental setup on a robotic arm}
\label{sec:experiment}

We test the model on the multisensory UR-5 arm of robot TOMM \cite{dean2017tomm} as depicted in Fig. \ref{fig:setup}. Although the methodology is thought for robots difficult to calibrate with imprecise sensors, we use this platform as a proof of concept as we can easily compare with the ground truth values. Without loss of generality, the body (latent variables) is defined as the joint angles and its perception from multiple modalities: (1) the proprioceptive input data is three joint angles with Gaussian added noise (shoulder$_1$, shoulder$_2$ and elbow - Fig. \ref{fig:data:a1}); (2) the visual input is a rgb camera mounted on the head of the robot with $640\times480$ pixels definition; and (3) the tactile input is generated by multimodal skin cells distributed on the arm \cite{mittendorfer2011humanoid}. 

\subsection{Learning $g(x)$ from visual and proprioceptive data}
\begin{figure}[!hbtp]
\centering
\subfigure[Data recorded example (joint angles + noise, end-effector visual, end-effector cartesian) and schematic picture of the 3-DOF.]{\includegraphics[width=0.90\columnwidth, height=130px]{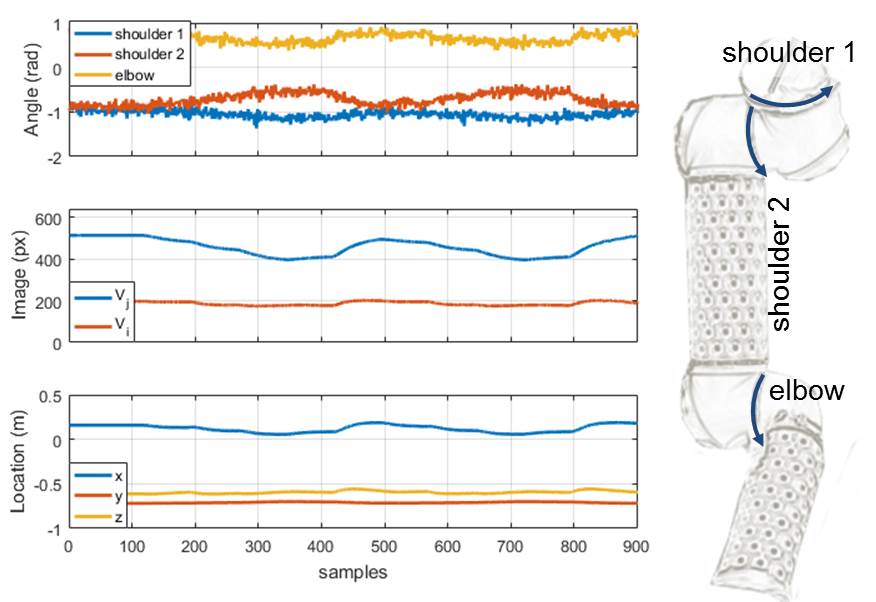} 
\label{fig:data:a1}}\\
\subfigure[Learnt $g_v(x)$ for visual horizontal displacement]{\includegraphics[width=0.90\columnwidth, height=85px]{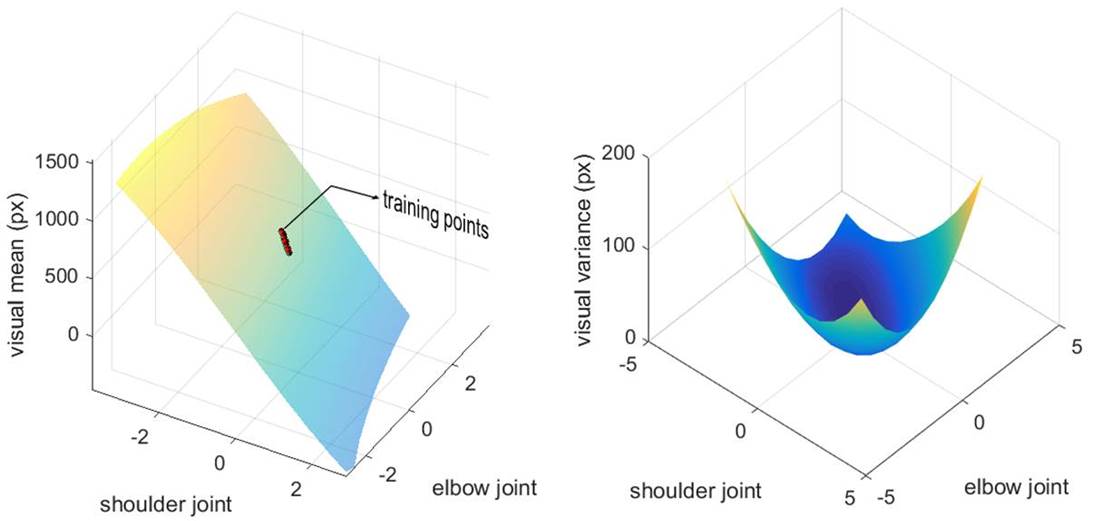}
\label{fig:data:a2}}\\
\subfigure[Skin proximity data]{\includegraphics[width=0.95\columnwidth, height=65px]{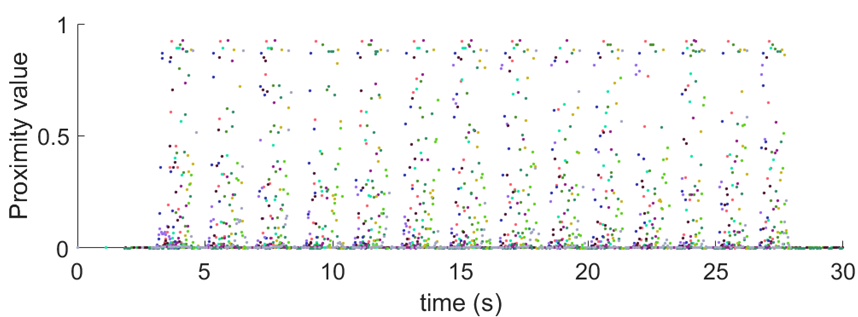}
\label{fig:data:a3}}\\
\subfigure[Tactile (left) and visual (right) event trajectories]{\includegraphics[width=0.80\columnwidth,height=80px]{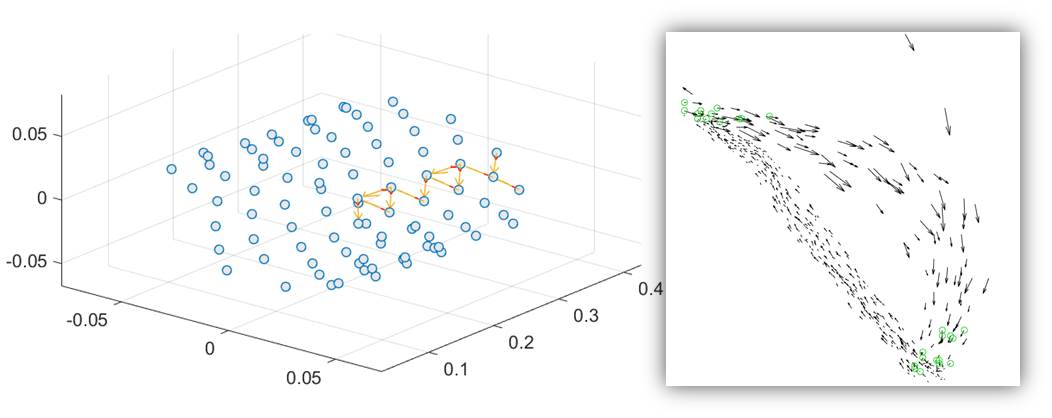} 
\label{fig:sync:a1}}
\caption{Collected data. (a) Joints, visual and ground truth information of the end-effector. (b) Example of the mean and the variance computed by the GP, which describes the visual horizontal displacement depending on two joints. (c) 30 seconds of raw proximity sensor information of 117 forearm skin cells. (d) Touch patterns extracted from tactile and visual sources.}
\label{fig:data}
\end{figure}

In order to learn the sensory forward/observation model we programmed random trajectories in the joint space that resemble to horizontal displacements of the arm. Figure \ref{fig:data:a1} shows the data extracted: noisy joint angles and visual location of the end-effector, obtained by colour segmentation. To learn the visual forward model $s_v = g_v(x)$, each sample is defined as the input joint angles sensor values $x= (x_1, x_2, x_3)$ and the output $s_v = (i,j)$ pixel coordinates. As an example, Fig. \ref{fig:data:a2} shows the learnt visual forward model by GP regression with 46 samples (red dots). The horizontal displacement mean (in pixels) with respect to two joint angles and the variance.

\subsection{Extracting visuo-tactile data}
We use proximity sensing information from the infrarred sensors located in every skin cell to discern when the arm is being touched. The infrarred sensor outputs a filtered signal $\in (0,1)$. The likelihood of a cell being touched is given by the following function (Eq. \ref{eq:touch}) $a_1 e^{-b_1\sum(s_v-s_o)^2}$, where $a_1 = 0.001$ and $b_1= 1$. The parameters have been obtained by fitting the function to the distance-sensor output measurements. Figure \ref{fig:data:a3} shows the raw skin proximity sensing data during the experiment (each colour represents the 117 different skin cells). From the other's hand visual trajectory and the skin proximity activation we compute the level of synchrony between the two patterns (Fig. \ref{fig:sync:a1}). Timings for tactile stimuli are obtained by setting a threshold over the proximity value: prox $> 0.7 \rightarrow$ activation. Timings for other's trajectory events are obtained through the velocity components. Detected initial and ending position of the visual touching is depicted in Fig. \ref{fig:sync:a1} (right, green circles).

\section{Results}
\label{sec:results}
For comparison purposes, all experiments parameters are set fixed values. $g_v(x)$ learning hyperparameters: signal variance $\sigma_n = \exp(0.02)$ and kernel length scale $l=\exp(0.1)$. The integration step is $\Delta_t= 0.05$ ($20Hz$) and error variances are $\sigma_x \in \mathcal{R}^3 = [1,1,1]$, $\sigma_p \in \mathcal{R}^3 = [1,1,1]$, $\sigma_v \in \mathcal{Z}^2= [5,5]$. Finally, the learning rate of $\mu_x$ is $\lambda = 1$.

\begin{figure*}[!t]
\centering
\subfigure[Joint angles estimation]{\includegraphics[width=0.32\textwidth, height=120px]{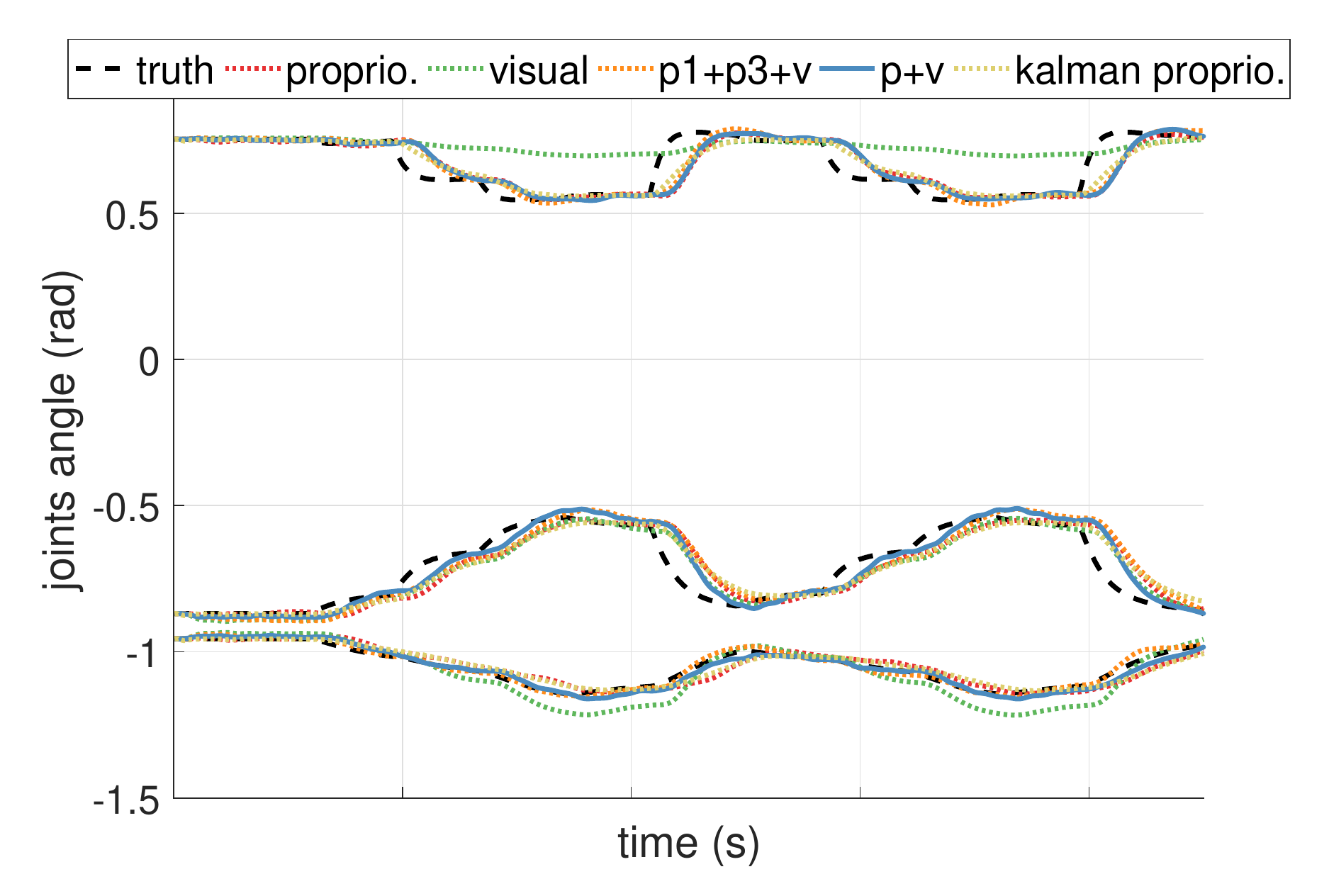} 
\label{fig:results1:a}}
\subfigure[Estimation error]{\includegraphics[width=0.32\textwidth]{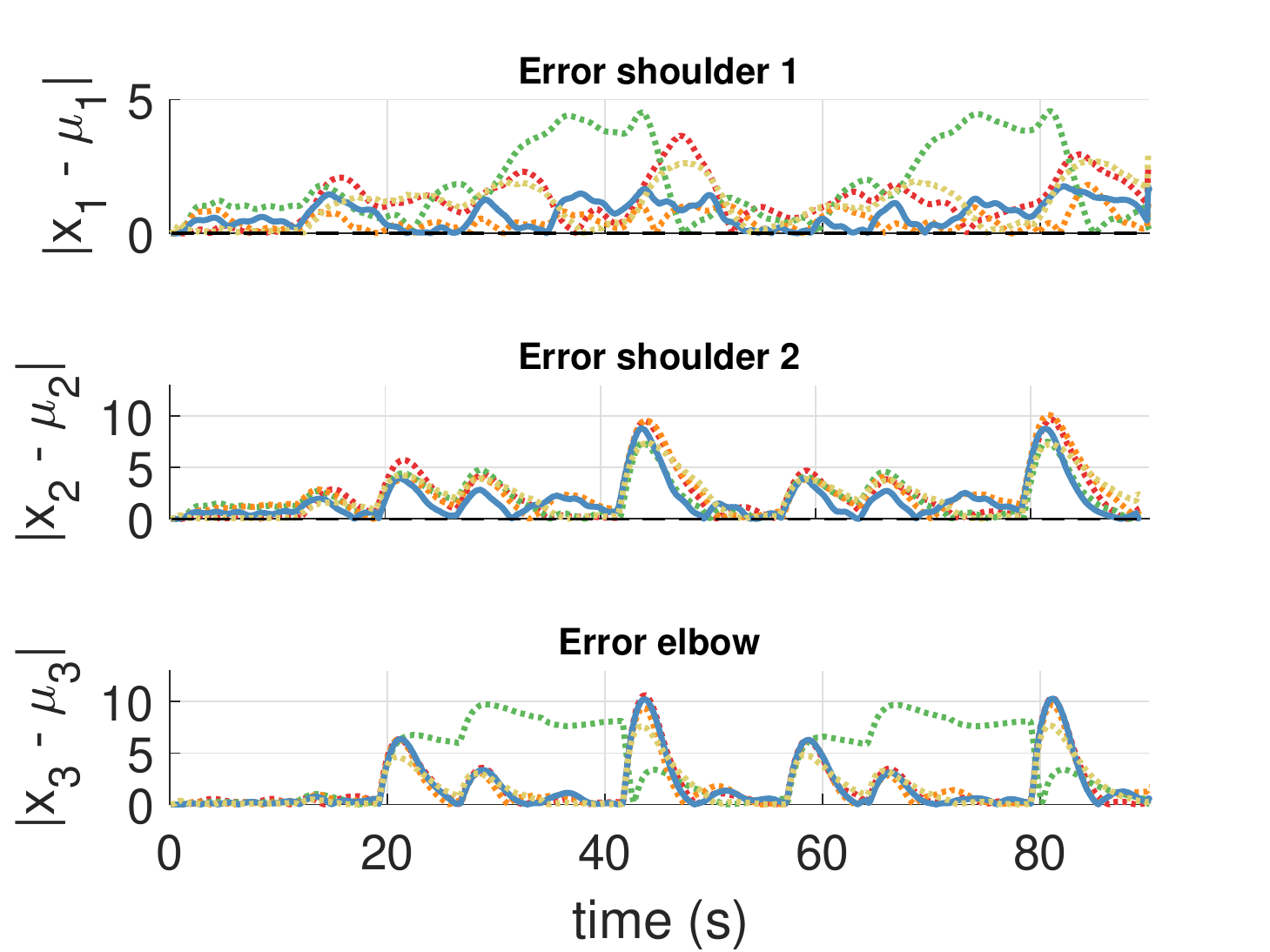} 
\label{fig:results1:b}}
\subfigure[Error dynamics]{\includegraphics[width=0.32\textwidth]{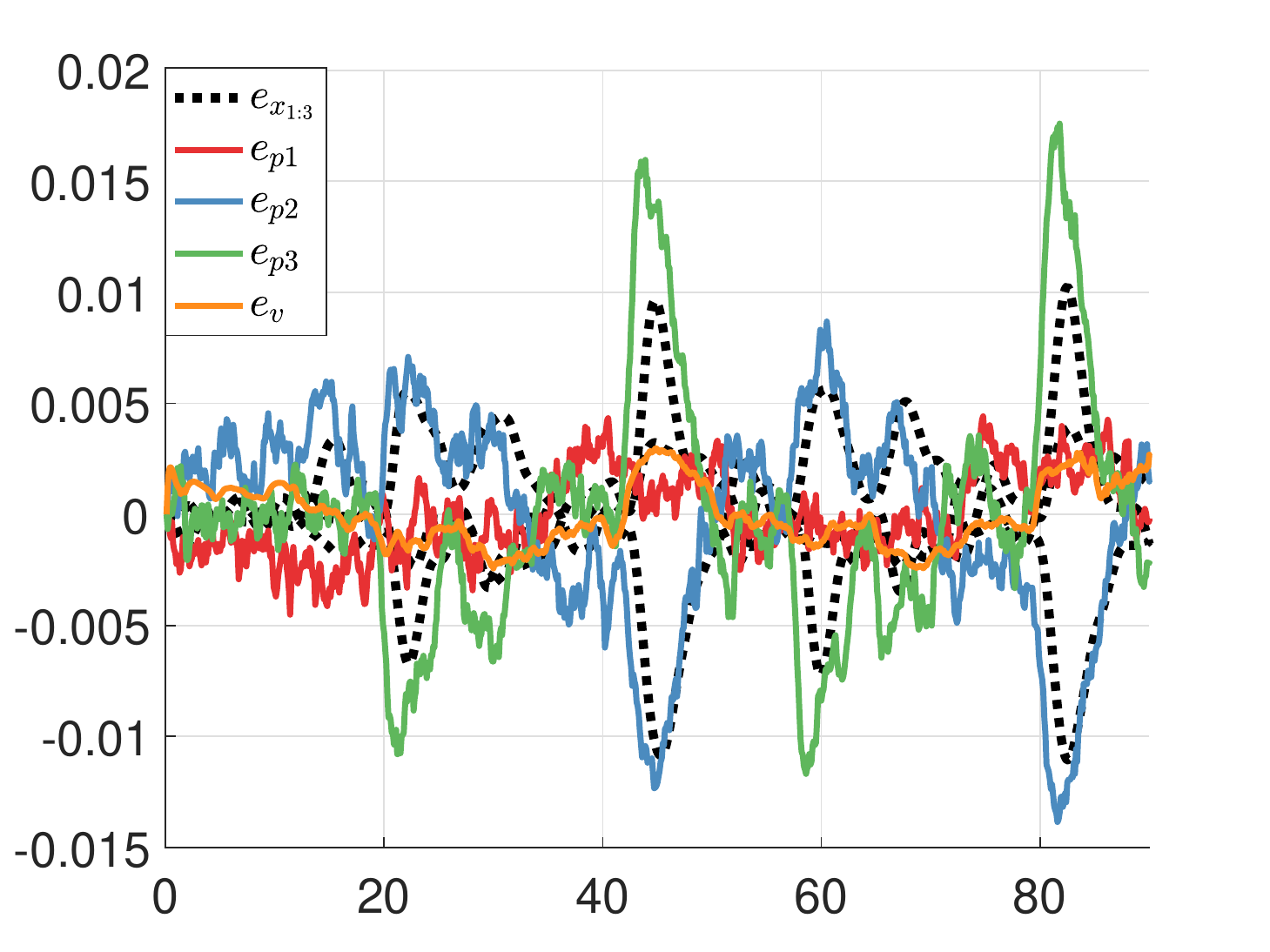} 
\label{fig:results1:c}}
\caption{Body estimation (joint angles) and error comparison. Proposed algorithm tested with different sensor contributions (only visual; visual + two joint sensors; and visual + proprioception) and compared with a Kalman filter just using the joint measurements - see text for details.}
\label{fig:results1}
\end{figure*}
\subsection{Robust multisensory integration}
We present three different experiments to study visual and proprioceptive body estimation. The first one, described in Fig. \ref{fig:results1} shows the proposed body estimation algorithm while deploying a similar trajectory as presented in Fig. \ref{fig:data:a1}. We analyse the error between the estimated body configuration and the ground truth joint angles for different sensor contributions. The algorithm is able to correctly estimate the joint angles but presents slow dynamics when big changes occur, due to the static nature of the generative model used. It also shows that with only visual input it is not able to estimate the elbow angle. This happens because learning trajectory was set to not provide information about the elbow. However, we can see how combining visual information and two joint sensors ($p1+p3+v$), reduces the estimation error. This shows the ability of the proposed method to deal with missing information. We have also validated the method against an standard Kalman filter \cite{besada2012localization} with only the joint angles as input (proprioception), process noise covariance $\sim diag(0.001)$, same measurement noise as the proposed approach and static transition model for fair comparison (yellow dotted-line). As expected, the error and behaviour is practically equivalent to the proprioception version of the proposed approach (red dotted-line).

\begin{figure}[!hbtp]
\centering
\subfigure[Body estimation with non-linear proprioception]{\includegraphics[width=0.46\columnwidth, height=90px]{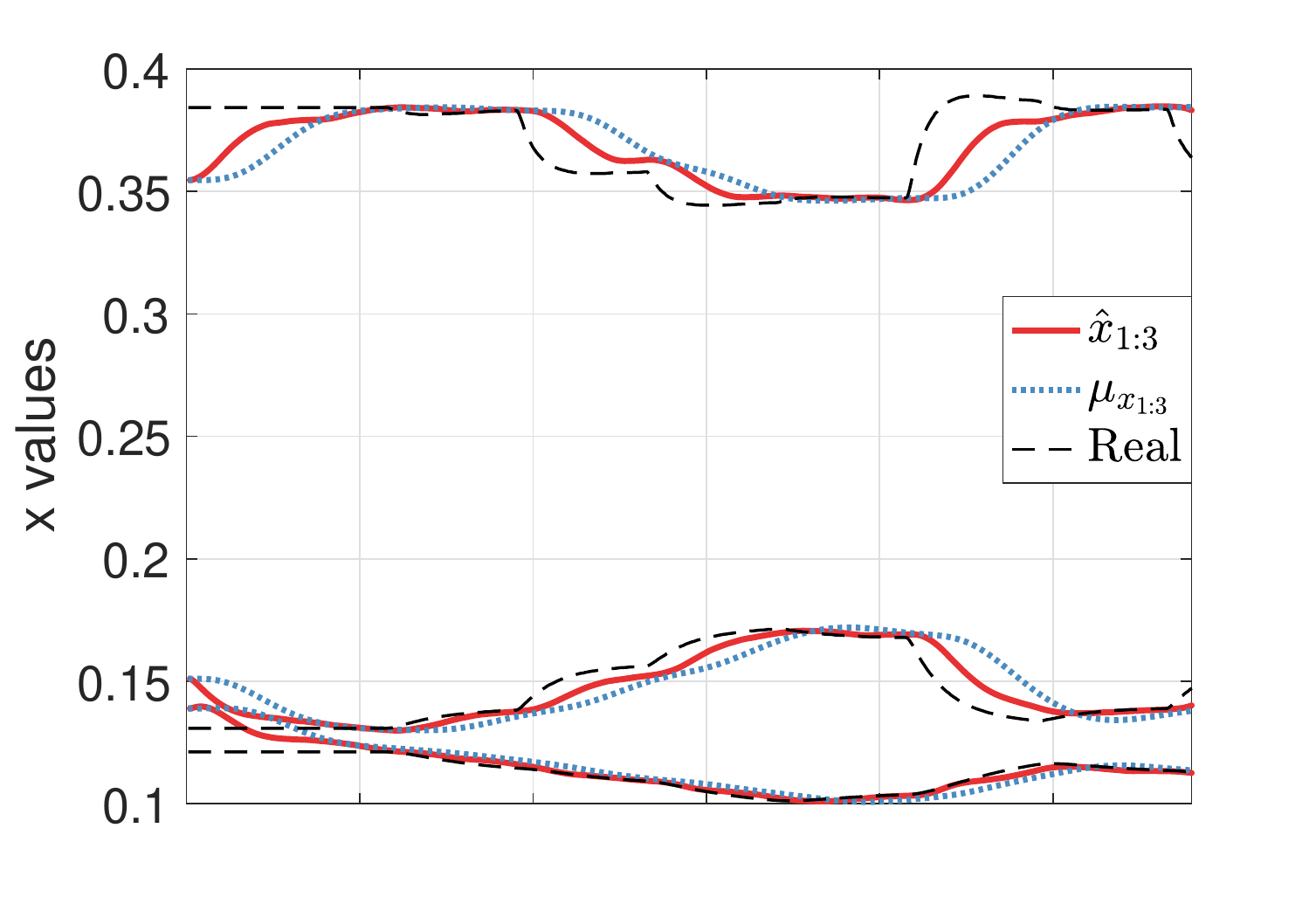} 
\label{fig:results2:a1}}
\subfigure[Damaged proprioception compesated with visual sensing]{\includegraphics[width=0.45\columnwidth,height=90px]{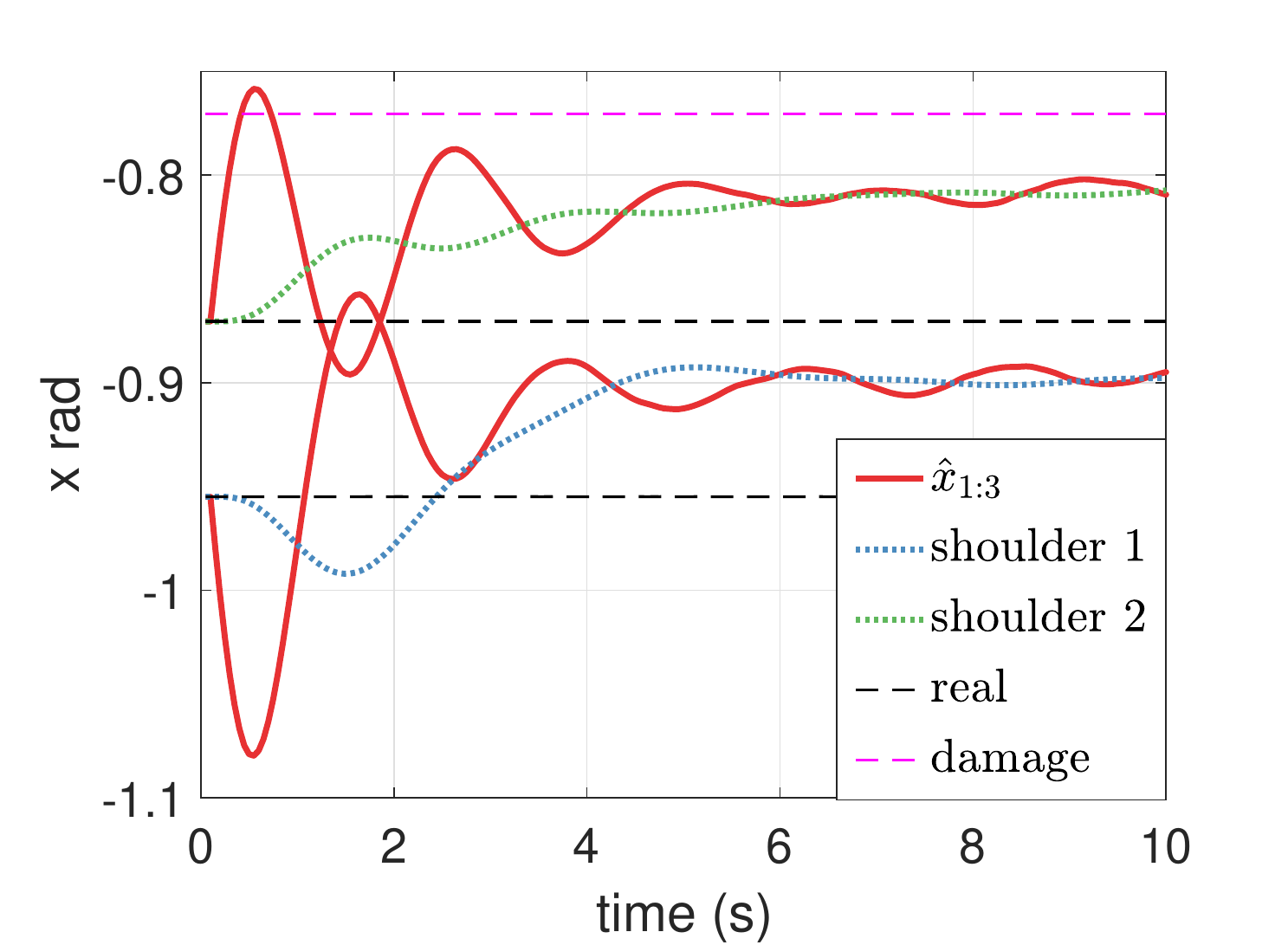} 
\label{fig:results2:a2}}
\caption{Non-linear and damaged proprioception test. (a) The generative model of the proprioceptive sensing is quadratic plus Gaussian noise $x^2 + z_p$. (a) Body configuration estimation vs real in the \textit{proprioceptive values space}. The initial prior joint angles $\mu_x = [-0.8, 0.70, 0.60]$ differs from the real one $x = [-0.9550, -0.8692, 0.7532]$. (b) Multimodal body inference with biased $shoulder_1$ sensor.}
\label{fig:results2}
\end{figure}

In second experiment, presented in Fig. \ref{fig:results2:a1}, we test the model with non-linear proprioceptive sensors: $x^2$. The body configuration values plotted are in the sensor space. We have initialized the robot body belief with a wrong configuration. On the first 5 seconds, the plot shows how the system converges to the ``embodied" configuration and then the arm starts moving. The estimation reaction time is slightly slower than previous experiment. Furthermore, we observe an interesting effect. The joint angles vary from $-\pi$ to $\pi$, but with the function $x^2$ the robot cannot distinguish between positive and negative angles. Thus, when inverting the sign of one joint the robot thinks that it is in the right configuration but it is not.

The last experiment, depicted in Fig. \ref{fig:results2:a2}, we study how the model deal with damaged or uncalibrated sensors. After the visual learning stage, we have added a drift error to shoulder$_1$ proprioceptive sensor. The visual prediction error should correct this anomaly. The plot shows how the system nicely reduces proprioceptive drift in shoulder$_1$. However, it induces a wrong bias on shoulder$_2$. Thus, although visual information, with the current $g_v(x)$ learning, evidences a coupling between $x_1$ and $x_2$, visual correction has appeared.

\subsection{Adaptation with visual, proprioceptive and tactile sensors}
\begin{figure}[!hbtp]
\centering
\subfigure[Body estimation refinement with biased prior]{\includegraphics[width=0.95\columnwidth, height=80px]{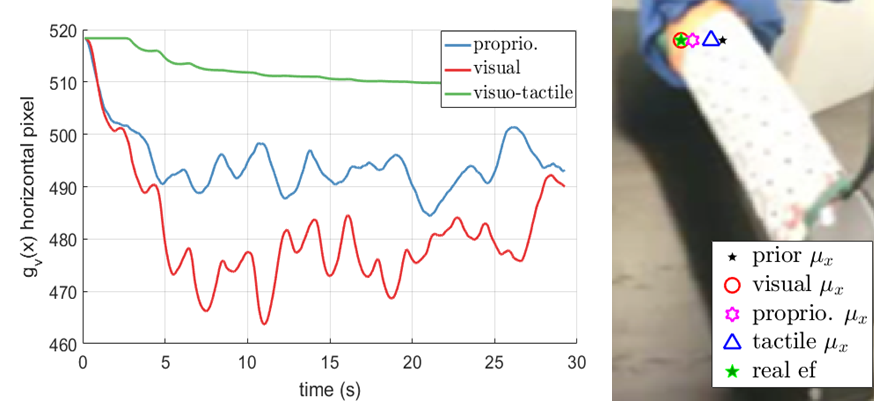} 
\label{fig:resultsrh:a1}}
\subfigure[Joint angles estimation with wrong visuo-tactile sensory input]{\includegraphics[width=0.93\columnwidth, height=80px]{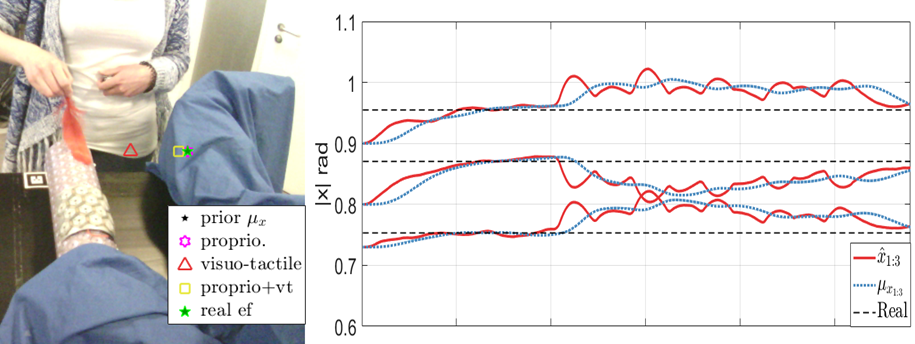} 
\label{fig:resultsrh:a2}}
\caption{Adaptation analysis. (a) Joint angles estimation using different sensor modalities. (c) Body inference with visuo-tactile perturbation: body latent variables (blue dotted line) and ground truth (black dotted line).}
\label{fig:resultsrh}
\end{figure}

We further test the proposed model adaptation with proprioceptive and visuo-tactile stimulation. Fig. \ref{fig:resultsrh:a1} describes body estimation refinement depending on different sensor modalities. Every sensor or feature contributes independently to improve the robot arm localization. In essence, the method provides scalable data association, e.g., the robot can learn more than one visual feature and incorporate them into the predictive error formulation as an additive term. Besides, Fig. \ref{fig:resultsrh:a2} experiment shows the potential of the proposed method to adapt its body inference to incoherent new situations as a human will do. We have introduced a strong perturbation on the visuo-tactile input inspired by the rubber-hand illusion experiments in humans~\cite{hinz2018drifting}. The new visual location induced by synchronous tactile stimulation makes the robot to infer the most plausible situation given the sensory information, which in this case is to drift the location of the arm towards the new location. In the first 5 seconds, there is no tactile stimulation and the estimation is refined to ground truth (black dotted line). Then we inject visuo-tactile stimulation while other agent is pretending to touch another location. When it becomes synchronous a horizontal drift appears and the inferred body configuration is altered. 

\subsection{A note on scalability}

The learning using Gaussian process regression has a computational complexity of $\mathcal{O}(n^3)$ and the prediction of the sensor forward model depends on the covariance kernel complexity $\mathcal{O}(\text{kernel})$. For $N$ independent sensor contributions, $M$ internal variables and $N_S$ samples, the prediction of forward models is $\sim \mathcal{O}(N \times N_S)$. Finally, the free-energy optimization is $\mathcal{O}(N+M)$ using Euler integration method.

\section{Discussion: I sense, therefore I am?}
\label{sec:discussion}
We have stressed that robot body estimation can be computed just by means of sensory information. Every sensing modality or feature, when available, contributes to the final body estimation through the prediction error and the variance of each error describes the precision of every sensor with respect to body internal variables. For instance, outside of the field of view proprioceptive and tactile sensors define the arm configuration. When the arm appears in the visual field, other features are included into the inference. We have also shown that when the robot has a broken proprioceptive sensor it can rely on visual features to complete the lack of information. Finally,  we have underscored embodiment showing how the sensor function influences body estimation. Hence, we have defined adaptive body learning and estimation as providing the most plausible solution according to the current information available from the sensors. As a collateral effect, the model has been showed to be prone to visuo-tactile illusions, something that has been also evidenced in humans.

Nevertheless, we have only focused on passive perception and omitted deliberatively the generative model of the body dynamics. Moreover, where is the action? We have not considered it in the model, something core for interacting with the body. In order to obtain the full construct, which properly reduces the KL-divergence between the robot belief and the posterior probability of the body configuration given the sensors, we need to include the robot dynamics. However, this is a hard task from the learning perspective. The advantage with this approach is that we only need an approximation of the dynamics because free-energy minimization should solve the discrepancy. With the full construct we expect to improve prediction accuracy and to incorporate the action into the body estimation framework.


\section{Conclusion}
\label{sec:conclusion}
We have presented an adaptive robot body learning and estimation algorithm based on predictive processing, able to integrate information from visual, proprioceptive and tactile sensors. The robot independently learns the sensor forward generative functions and then it use them to refine its body estimation by a free-energy minimization scheme.

The model has been tested on a robot with a standard industrial arm to facilitate ground truth comparison. Results have shown how the model deals with missing and noisy sensory information, reducing the effect of sensor failures. The algorithm has also displayed adaptability to wrong body prior initialization and unexpected situations. In addition, we have shown how other's touch can refine body robot estimation, opening interesting questions about improved localization and mapping by means of tactile interaction. Altogether reflects the potential of the proposed approach for complex robots, where estimating body location is a hard task and a requirement for safe interaction. 

\addcontentsline{toc}{section}{References}
\bibliographystyle{IEEEtran}

\bibliography{pl,selfception}

\end{document}